\documentclass[10pt,twocolumn,letterpaper]{article}

\usepackage[pagenumbers]{wacv}
\usepackage{graphicx}
\usepackage{booktabs}
\usepackage{multirow}
\usepackage{color}
\usepackage{soul}
\usepackage{comment}
\usepackage{subcaption}
\usepackage{colortbl}
\usepackage{array}
\usepackage{makecell}
\usepackage{adjustbox}
\usepackage[table]{xcolor}
\usepackage{multirow}

\usepackage{adjustbox}
\usepackage[table,xcdraw]{xcolor}
\usepackage{algorithm}
\usepackage{algpseudocode}
\usepackage[justification=centering]{caption}
\usepackage{float}
\usepackage{stfloats} 
\newcommand{\methodName}{SemPurify\xspace}

\definecolor{wacvblue}{rgb}{0.21,0.49,0.74}
\usepackage[pagebackref,breaklinks,colorlinks,allcolors=wacvblue]{hyperref}

\def\wacvPaperID{1305} 
\def\confName{WACV}
\def\confYear{2027}

\title{SemPurify: Semantics-Aware Data Purification for Black-box Backdoor Defense}

\author{
Jiawei Yu$^{1}$ \qquad Syed Irfan Ali Meerza$^{2}$ \qquad Yi Wu$^{3}$\\
Amir Sadovnik$^{4}$ \qquad Jian Liu$^{1}$\\[0.6em]
$^{1}$University of Georgia \qquad $^{2}$Virginia Commonwealth University\\
$^{3}$University of Oklahoma \qquad $^{4}$Oak Ridge National Laboratory
}

\begin{document}
\maketitle
\begin{abstract}
Backdoor attacks have emerged as a critical security threat in machine learning, where adversaries embed triggers within poisoned training samples to manipulate a model's predictions on targeted inputs while leaving its general behavior unaffected. This stealthy nature makes backdoor attacks difficult to detect, as the model functions normally until activated by a specific malicious trigger. Current purification-based defenses use image transformations (e.g., blurring) to disrupt trigger patterns but often degrade image quality. Even with advanced restoration techniques like diffusion models, {these approaches can still alter visual or semantic information}, reducing utility for downstream tasks. To address this, we propose the first semantics-aware data purification framework for black-box backdoor defense, \methodName, which {explicitly incorporates} multi-level semantics (i.e., low-level semantic embeddings and high-level image tags) within a text-to-image (T2I) framework {to guide image restoration during purification}. Using this semantic information, the T2I model subsequently refines the transformed samples, {helping preserve visual details and semantic information during restoration}. Extensive experiments spanning multiple backdoor scenarios, i.e., {classification models, CLIP-style encoders, and generative models}, demonstrate that \methodName reliably removes backdoors while {effectively preserving semantic information for downstream tasks}.
\end{abstract}

\vspace{-8mm}
\section{Introduction}
\vspace{-2mm}

Backdoor attacks~\cite{blended, gao2020backdoor} pose a covert yet severe threat to machine learning systems.
In such attacks, adversaries embed hidden triggers into training data so that the trained model behaves normally on clean inputs but produces attacker-specified outputs when the trigger is present.
At inference time, these triggers cause the model to follow attacker-specified behaviors: targeted misclassification in classification models; targeted prediction or malicious prompt generation in CLIP-style encoders~\cite{bai2024badclip, liang2024badclip, liang2025vl}; and attacker-directed content synthesis in generative models (e.g.,~\cite{rawat2022devil, I2IUAP}).

To address the concerns caused by backdoor attacks, researchers have developed a variety of defense methods. Many existing approaches~\cite{zeng2021rethinking,guo2021aeva,dong2021black,li2021anti,guo2023scale, hou2024ibd, zhang2025test,tun2025ocage} focus on detecting and filtering poisoned samples before training. However, simply removing such samples {can lead to a loss of critical training data}.
Other lines of defense rely on additional assumptions, such as prior knowledge of clean samples~\cite{li2021neural, zhu2023enhancing,li2023reconstructive,wei2025backdoor} or access to the model~\cite{li2021anti,chen2025refine,hou2025flare}. While effective under ideal conditions, these methods {can become inapplicable when such information is unavailable}.
To overcome these limitations, recent works have explored purification-based defenses~\cite{Physicalba,ZIP,zhou2024dataelixir,yang2024sampdetox, li2025bdfirewall} which aim to remove backdoor triggers directly from data without requiring model access. These methods are particularly attractive in black-box scenarios, where defenders cannot inspect the model or access clean samples. However, existing purification approaches still suffer from important limitations. Some methods (e.g.,~\cite{Physicalba}) provide limited protection against sophisticated backdoor attacks, while others (e.g.,~\cite{ZIP,zhou2024dataelixir}) {can alter visual details or semantic information during purification}, which can adversely affect the performance of downstream tasks.

The challenges of backdoor attack and current defense limitations raise a key question: \textit{Can we eliminate backdoor triggers while {preserving semantic information} in a fully black-box setting, without access to clean samples or prior knowledge of the model?}
In this paper, we propose \methodName, {a semantics-aware, black-box data purification framework} designed to disrupt backdoor triggers, if present, while preserving {semantic information}. Unlike some of the prior methods, \methodName operates without access to model parameters, the training dataset, or knowledge of which samples are clean or poisoned.

Specifically, to retain semantic details throughout the data purification process, \methodName first extracts multi-level semantics, including both low-level semantic embeddings and high-level image tags, from the data (either poisoned or clean samples) using an image tagging model, i.e., Recognize Anything Model (RAM)~\cite{ram}. {These two forms of semantic information are subsequently used to guide the restoration process.}

To disrupt hidden poisoned triggers, we apply linear transformations~\cite{ZIP,li2020rethinking} (e.g., blurring); however, these transformations can also impact the data's semantic content, leading to a loss of important visual details. To address this, we leverage the extracted semantic embeddings and image tags to guide data restoration through a pre-trained text-to-image (T2I) model, such as Stable Diffusion~\cite{SD}, conditioned with ControlNet~\cite{zhang2023adding} {for structural guidance}.
{By explicitly incorporating multi-level semantic guidance into the restoration process, \methodName preserves semantic information while effectively suppressing backdoor effects.} 

Our main contributions are summarized as follows:

\begin{itemize}
     \item {We propose the first semantics-aware black-box data purification framework that explicitly incorporates multi-level semantic information into the restoration process to remove backdoor triggers while preserving semantic information.}

     \item We explore the benefits of combining low-level semantic embeddings with high-level semantic descriptions (i.e., image tags) to guide the purification process and empirically demonstrate that {the two forms of semantic information provide complementary guidance for trigger removal and semantic preservation}.

     \item Extensive experiments across multiple datasets, diverse backdoor attacks, and state-of-the-art purification defenses for classification models, CLIP-style encoders, and image-to-image generative models, {demonstrate the effectiveness of \methodName in mitigating backdoor attacks while preserving semantic information across all three settings.}
\end{itemize}

\vspace{-4mm}
\section{Related Work}
\vspace{-2mm}
\textbf{Backdoor Attacks.}
Backdoor attacks inject hidden triggers into a model via a poisoned training set, causing inputs with the trigger to be misclassified to an adversary-chosen label during inference. {Backdoor attacks in the image domain include patch-based, perturbation-based, and warping-based attacks.} Patch-based attacks embed visible pixel patches~\cite{badnet,Physicalba,turner2019label,zeng2023narcissus,tang2021demon,qi2023revisiting,nguyen2020input,qi2022towards} onto images, while perturbation-based attacks introduce imperceptible noise~\cite{blended,li2021invisible,gao2024backdoor, wang2022bppattack} across the image. {Warping-based attacks introduce geometric distortions to the input, as exemplified by WaNet~\cite{nguyen2021wanet}.}
For instance, BadNet~\cite{badnet} exemplifies patch-based backdoor attacks, leveraging pixel patterns for high success rates. Building on BadNet, PhysicalBA~\cite{Physicalba} improves robustness by incorporating physical transformations. Blended~\cite{blended} represents perturbation-based attacks, using subtle patterns that remain effective when replaced with semantically meaningful triggers.

\noindent \textbf{Defenses Against Backdoor Attacks.}
In this paper, we focus on inference-time defenses, which are generally black-box and can be broadly categorized into two classes: detection-based and purification-based. While numerous research efforts have focused on defending against backdoor attacks through detecting poisoned samples~\cite{zeng2021rethinking,guo2021aeva,dong2021black,li2021anti,guo2023scale, hou2024ibd, zhang2025test,tun2025ocage}, these approaches do not remove triggers from the poisoned samples, {limiting the subsequent use of these samples during inference}. Consequently, these methods are not applicable when critical poisoned samples are required for downstream models.

In contrast, purification-based approaches aim to remove backdoor triggers from input samples. For example, Li \textit{et al.} \cite{Physicalba} propose a transformation-based method using operations such as flipping or scaling to reduce the impact of triggers. However, {such transformations can alter visual details and semantic information in the input}. Chen \textit{et al.}~\cite{chen2025refine} introduce a U-Net-based transformation module to eliminate backdoor patterns. However, this method heavily relies on an identically distributed benign dataset, which is a restrictive condition rarely met in practical black-box scenarios. May \textit{et al.}~\cite{may2023salient} and Sun \textit{et al.}~\cite{sun2023mask} employ masking techniques to identify and mask backdoor patterns. However, both struggle when backdoor patterns are distributed throughout the image (e.g., perturbation-based attacks).

Some purification methods~\cite{huang2022backdoor,li2021anti, hou2025flare} focus on purifying training sets to produce clean models. However, their reliance on ground-truth labels and model architectures renders them impractical in black-box scenarios. In parallel, other methods perform purification during the inference stage. ZIP~\cite{ZIP} uses a pre-trained diffusion model to restore semantics lost during purification, but {the restoration process can still alter fine-grained visual details}. Similarly, DataElixir~\cite{zhou2024dataelixir}, adaptable to both training and inference stages, integrates label detection with diffusion-based purification to eliminate backdoor patterns. However, {its purification process can alter visual details and semantic information while also introducing additional computational overhead through the detection mechanism}. SampDetox~\cite{yang2024sampdetox} introduces a two-stage noise diffusion mechanism to identify high-visibility triggers via structural similarity (SSIM) and selectively apply intensive noise for localized purification. Nevertheless, {the stronger perturbation applied in its second stage can also alter semantic information during restoration}. BDFirewall~\cite{li2025bdfirewall} introduces a three-tier detection mechanism to perform different purification strategies on poisoned samples. However, its method still relies on a clean subset to calibrate detection thresholds, making it impractical in strict black-box scenarios, where accessing task-specific clean data is impossible.

Unlike prior efforts, \methodName explicitly extracts and utilizes multi-level semantics from samples {to guide the restoration process during purification}. {This explicit semantic guidance helps preserve semantic information while mitigating backdoor effects.}



\vspace{-2mm}
\section{Threat Model and Multi-level Semantics}
\vspace{-2mm}
\subsection{Threat Model}
\vspace{-2mm}

\methodName can defend against backdoor attacks across various backdoor scenarios including classification, CLIP-style, and generative models.

The objective of an image classification model, $F(x)$, is to classify the input image $x$ to its ground-truth label {$y_{\text{true}}$}.
For a CLIP-style encoder, $C(x)$, {we consider the few-shot image classification, where the goal is similarly to classify the input image $x$ to its ground-truth label $y_{\text{true}}$.} For a generative model (e.g., image-to-image model), the goal is to train a model $G(x)$ that transforms the input image $x$ into a target image {$x_{\text{target}}$} for various tasks, such as resolution enhancement. Backdoor attacks work by injecting triggers into the training data, resulting in poisoned models, denoted $F_p(\cdot)$, $C_p(\cdot)$, and $G_p(\cdot)$. During inference, the attacker poisons the test sample $x$ by adding a trigger $p$, creating a poisoned sample $x_p$.
The attacker seeks to force the poisoned classifier $F_p(\cdot)$ to predict a targeted label $y_{\text{target}}$ for a poisoned input $x_p$, instead of its true label $y_{\text{true}}$. Similarly, a poisoned CLIP-style encoder $C_p(\cdot)$ is induced to map $x_p$ toward the targeted label. For generative models $G_p(\cdot)$, the attacker aims to inject specific patterns into the output, such as generating an unintended image.

\textbf{Defender's Capability \& Objectives.} Our threat model assumes a challenging black-box scenario where the defender has no prior knowledge of the attack settings, including access to the model's parameters, training data, or information on whether an input image used during inference is poisoned or clean. Consequently, \methodName is designed to purify both clean and potentially poisoned test samples, enhancing resilience against backdoor attacks in unknown environments. {We assume access to an independently pre-trained and trusted image-tagging model that is not compromised by the backdoor attack.}
Additionally, \methodName is guided by two primary objectives:

\noindent \textbf{(1) Trigger Removal:} Effectively eliminate any embedded triggers in poisoned samples $x_p$, enabling accurate model behavior during inference. This objective ensures that the poisoned classification model yields correct predictions and the generative model produces the intended, unaltered patterns, mitigating the impact of backdoor triggers.

\noindent \textbf{{(2) Semantic Preservation:}} {Preserve the semantic information of data samples so that purification retains task-relevant content for downstream tasks. This objective helps maintain downstream performance on both clean and poisoned inputs.}

\begin{figure}[t]
    \centering
    \includegraphics[width=1\linewidth]{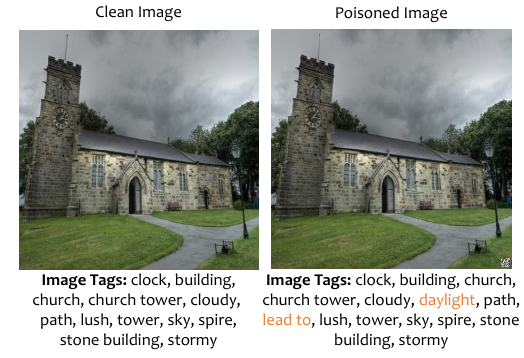}
    \vspace{-7mm}
    \caption{{Illustration of semantic descriptions extracted by the tagging model RAM~\cite{ram} from clean and backdoored inputs.} BadNet~\cite{badnet} is applied to cause the image labeled as ``church'' to be misclassified as ``dog''. While two tags (highlighted) are introduced, they remain contextually related to the original image, with none associated with the target label ``dog''.}
    \vspace{-6mm}
    \label{fig:tag_comparision}
\end{figure}

\vspace{-2mm}
\subsection{Multi-level Semantics Analysis}
\vspace{-2mm}
\label{subsec:multi-semantics}

In image understanding, multi-level semantics combine low-level and high-level semantic information. Low-level semantics, or semantic embeddings, are high-dimensional feature representations extracted from a pre-trained image encoder, capturing detailed aspects of an image. High-level semantics, or image tags, are generated by a tagging head and provide descriptive labels that summarize the main content of the image~\cite{tag2text, query2label}.

To enhance backdoor purification, \methodName combines semantic embeddings and image tags to preserve both fine-grained features and high-level semantic meanings. {The two forms of semantic information provide complementary guidance during restoration: semantic embeddings retain fine-grained visual information, while image tags describe prominent objects and scenes at a higher semantic level.} {Although different attacks perturb images in different ways, the semantic information extracted by RAM remains largely stable across the evaluated attacks. For example, the image tags extracted from the poisoned image in Fig.~\ref{fig:tag_comparision} do not contain the target label ``dog''.} {By combining these two levels of information, \methodName can preserve semantic information while reducing the influence of backdoor triggers during purification.}

\begin{figure*}[t]
    \centering
    \includegraphics[width=0.89\linewidth]{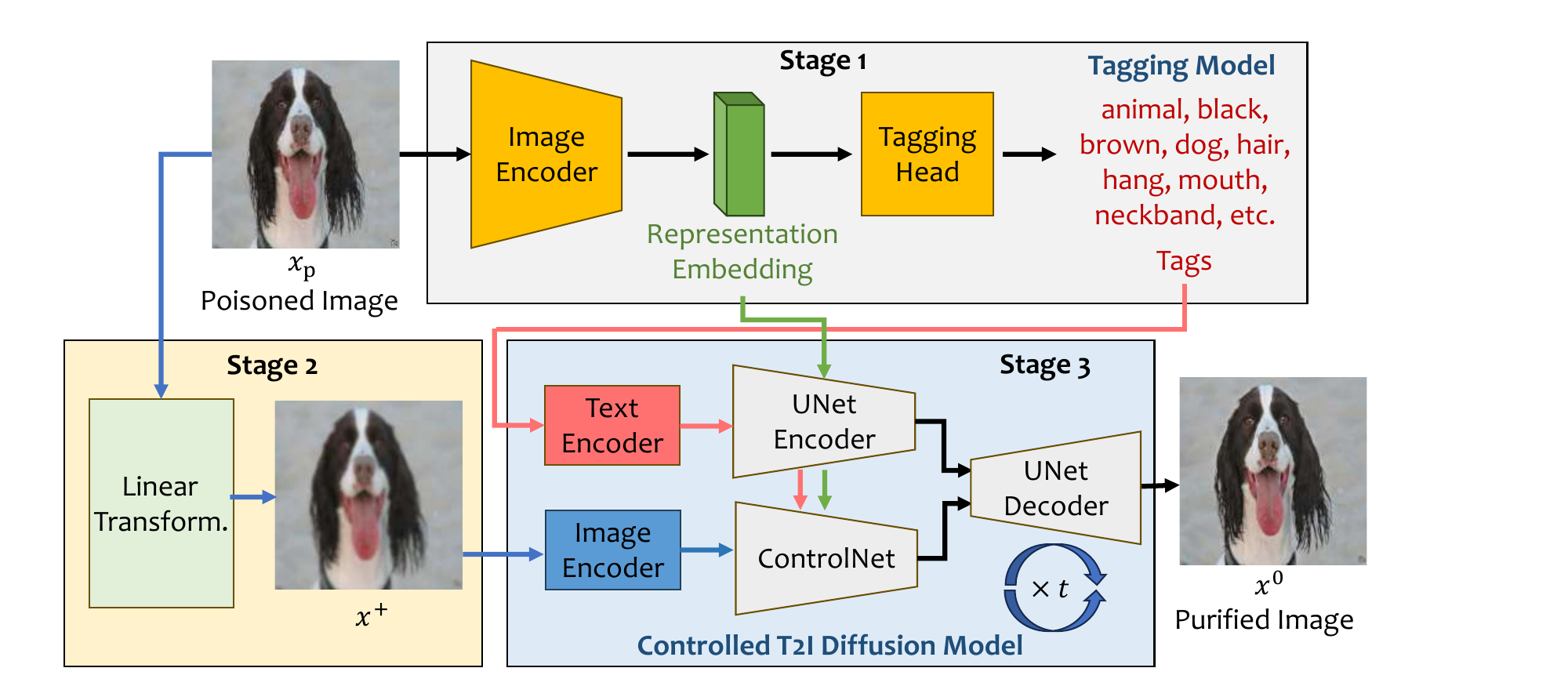}
    \vspace{-5mm}
    \caption{The overview of \methodName: In Stage 1, we use an image tagging model to extract multi-level semantics. In Stage 2, we use a linear transformation such as blurring to destruct the trigger patterns in the poisoned image $x_p$. In Stage 3, we use a conditional text-to-image model to restore the semantic information. This process ensures that $x^0$ retains the original semantic information from $x_p$ while effectively neutralizing any backdoor trigger patterns.}
    \vspace{-6mm}
    \label{Pipeline}
\end{figure*}

\vspace{-1mm}
\section{Methodology}
\vspace{-1mm}

\subsection{Overview of \methodName}

As shown in Fig.~\ref{Pipeline}, our purification begins with a potentially poisoned sample $x_p$ and passes it through a carefully designed pipeline to produce a purified sample $x^0$. Specifically, \methodName operates in three stages, each serving a distinct purpose to ensure both security against backdoor threats and preservation of essential content: \textit{(1) Multi-level Semantic Extraction}; \textit{(2) Linear Transformation}; and \textit{(3) Semantic Data Restoration}. Each stage will be described in the following sections.

\vspace{-2mm}
\subsection{Multi-level Semantic Extraction}
\vspace{-2mm}

In this stage, we extract multi-level semantics (i.e., semantic embeddings and image tags) from the data sample $x_p$. \methodName leverages the advanced image tagging model, Recognize Anything Model (RAM)~\cite{ram}, which employs a Swin Transformer~\cite{liu2021swin} as the image encoder to generate semantic embeddings and a two-layer transformer as the tagging head to produce high-level image tags. RAM's adoption is driven by its robust design, which effectively provides high-accuracy tags, outperforming image tagging models like Tag2Text~\cite{tag2text} and Query2Label~\cite{query2label}. These models rely on random learnable embeddings for tag decoding, whereas RAM leverages textual label queries derived from the 10,000 most frequent tags, processed by a pre-trained CLIP encoder~\cite{radford2021learning}. This embeds rich semantic context and facilitates the interaction between image representations and textual label queries for semantic extraction.

Formally, given an input potentially poisoned data sample $x_p$, the image encoder $\mathcal{E}$ generates semantic embeddings $S_e = \mathcal{E}(x_p)$, while the tagging head $\mathcal{H}$ produces image tags $S_t$ from the embeddings as $S_t = \mathcal{H}(S_e)$.

\vspace{-2mm}
\subsection{Semantic Data Restoration}
\vspace{-2mm}

In this stage, we employ a text-to-image (T2I) diffusion model to restore the semantic information in the transformed image $x^+$, which is obtained by applying a linear transformation $\mathcal{T}(\cdot)$ to the potentially poisoned sample $x_p$, i.e., $x^+ = \mathcal{T}(x_p)$. By leveraging the previously extracted multi-level semantics $S_e$ and $S_t$ to guide the reverse diffusion process, the purified image can recover semantic information lost during the transformation while maintaining the transformed image structure.

{To incorporate structural and multi-level semantic information into the restoration process, we augment the pre-trained T2I diffusion model with an image-conditioning encoder, ControlNet~\cite{zhang2023adding}, and representation cross-attention (RCA) modules~\cite{PASD}. Specifically, the transformed image $x^+$ is processed by the image-conditioning encoder and then provided to ControlNet as structural guidance. The high-level image tags $S_t$ are encoded by the pre-trained T2I text encoder and incorporated through the original text cross-attention (TCA) modules. In parallel, the semantic embeddings $S_e$ are incorporated through RCA modules placed after the TCA modules, providing feature-level semantic guidance. The conditional features generated by ControlNet are injected into the corresponding U-Net blocks during denoising. In this way, $x^+$ provides structural information, while $S_t$ and $S_e$ provide high-level and feature-level semantic information, respectively.}

{During training, the parameters of the pre-trained T2I backbone, including its VAE, text encoder, and original U-Net parameters, are kept frozen, while the image-conditioning encoder, ControlNet, and RCA modules are optimized. Given the latent representation $z_0$ of a target image, Gaussian noise $\epsilon$ is added according to a randomly sampled diffusion timestep $t$ to obtain the noisy latent $z_t$. The restoration model is optimized using the standard diffusion noise-prediction objective conditioned on the transformed image $x^+$ and the extracted semantic information $S_e$ and $S_t$:}
\begin{equation}
{
\mathcal{L}_{\mathrm{train}}
=
\mathbb{E}_{z_0,t,\epsilon}
\left[
\left\|
\epsilon -
\epsilon_{\theta}(z_t,t,x^+,S_e,S_t)
\right\|_2^2
\right],
}
\end{equation}
{where $\epsilon_{\theta}(\cdot)$ denotes the conditional denoising network. This objective trains the conditioning modules to inject the structural and semantic information into the frozen T2I backbone for semantic restoration.}

The reverse diffusion process, conditioned on the structural prior $x^+$ and the semantic embeddings $S_e$ and $S_t$, is formally defined as:
\begin{equation}
\setlength{\abovedisplayskip}{3pt}
\setlength{\belowdisplayskip}{3pt}
{z_{t-1} = \mathcal{D}(z_t, t, x^+, S_e, S_t) + \sigma_t \cdot \epsilon_t,}
\end{equation}
{where $\mathcal{D}(\cdot)$ denotes the denoising network equipped with the above conditioning modules. Here, $z_t$ is the noisy latent state at timestep $t$, $z_{t-1}$ is the denoised latent state, $\sigma_t$ denotes the standard deviation of the noise schedule, and $\epsilon_t \sim \mathcal{N}(0, I)$ is noise sampled from a standard Gaussian distribution.}

{Through this reverse process, the latent is progressively denoised by incorporating the structural information from $x^+$ and the semantic information from $S_e$ and $S_t$, and the resulting $z_0$ is decoded to produce the purified image $x^0$.}

\vspace{-2mm}
\section{Experiments}
\subsection{Experimental Setup}

\noindent\textbf{Backdoor Attacks, Datasets, and Baseline Methods.} We evaluate \methodName against backdoor attacks across three model types: {\underline{\textit{classification models}}, \underline{\textit{CLIP-style encoders}}, and \underline{\textit{generative models}}}. For {\underline{\textit{classification models}}}, we use ResNet-34~\cite{Resnet} as the backbone and evaluate on three widely used datasets: (1) CIFAR-10~\cite{Cifar10}; (2) GTSRB~\cite{GTSRB}; and (3) Imagenette~\cite{Imagenette}. We test against five representative backdoor attacks covering key categories in the image domain: patch-based, perturbation-based, and warping-based. Specifically, (1) BadNet~\cite{badnet} embeds a visible pixel patch into images as a standard patch-based attack; (2) PhysicalBA~\cite{Physicalba} optimizes patch-based attacks with physical transformations for greater robustness; (3) Blended~\cite{blended} blends invisible noise into images as a perturbation-based attack; (4) Semantic Blended~\cite{blended} enhances perturbation-based attacks by using semantically meaningful images as backdoor patterns instead of random noise; and (5) WaNet~\cite{nguyen2021wanet} applies imperceptible geometric distortions as a warping-based attack. These attacks capture the core mechanisms of their types and remain relevant in recent studies~\cite{ZIP, Physicalba, zhou2024dataelixir, chen2025refine}, enabling comprehensive and reproducible evaluation across diverse models and datasets.

For BadNet and PhysicalBA, we use a randomly generated $2 \times 2$ pixel backdoor pattern for images of size $32 \times 32$, and a randomly generated $9 \times 9$ pixel backdoor pattern for images of size $256 \times 256$.
For Blended and Semantic Blended, we set the blending rate to 0.2 and 0.1, respectively, to reduce detectability. In Semantic Blended, we use two cartoon images as backdoor patterns: a $32 \times 32$ ``Hello Kitty'' for CIFAR-10 and GTSRB, and a $256 \times 256$ ``Blue Elephant'' for Imagenette.
For WaNet, we set the grid size to match the input sample size.
{All attacks are configured with a 5\% poison rate.} To ensure fair comparison, we follow the settings from ZIP~\cite{ZIP}, with other configurations based on BackdoorBox~\cite{backdoorbox}. More details are in the supplementary material.

For \underline{\textit{CLIP-style encoders}}, we select vanilla CLIP~\cite{radford2021learning}, one of the most widely used pre-trained multimodal encoders, and evaluate it under the state-of-the-art backdoor attack BadCLIP~\cite{bai2024badclip}, which leverages prompt learning to embed invisible perturbations as backdoor triggers. Following the BadCLIP setup, we conduct experiments on two widely used classification datasets, Caltech101~\cite{fei2004learning} and OxfordPets~\cite{parkhi2012cats}.

For {\underline{\textit{generative models}}}, we focus on the super-resolution (SR) domain, selecting the EDSR model~\cite{EDSR} as a representative architecture. To evaluate the effectiveness of our method in this novel context, we {adopt} the image-to-image (I2I) backdoor attack~\cite{I2IUAP}, a state-of-the-art backdoor attack designed for generative models. Consistent with Jiang~\textit{et al.}~\cite{I2IUAP}, we use the BSD100~\cite{BSD100} and Set14~\cite{Set14} datasets for training and testing, respectively.

We compare \methodName with {five} representative state-of-the-art baselines: ShrinkPad~\cite{Physicalba}, {BDMAE~\cite{sun2023mask}}, ZIP~\cite{ZIP}, DataElixir~\cite{zhou2024dataelixir}, and SampDetox~\cite{yang2024sampdetox}. These baselines are widely recognized for their effectiveness in black-box purification settings. Specifically, ShrinkPad mitigates backdoor triggers through image transformation, {BDMAE localizes suspicious regions and restores them using a masked autoencoder,} while ZIP employs diffusion-based restoration to recover semantic information lost during purification. DataElixir combines a label detection module with diffusion-based purification to remove backdoor patterns. SampDetox {combines weak and intensive noise to remove backdoor patterns}.

To ensure a fair comparison under the black-box threat model, we configure all baselines to operate without access to model internals or task-specific clean datasets. In particular, DataElixir~\cite{zhou2024dataelixir} is applied only at inference time on the test set, avoiding any model retraining. Methods that require additional assumptions, such as access to clean datasets or model-specific information~\cite{chen2025refine}, are not included, as they fall outside our threat model.

{For the restoration network, we use a pre-trained restoration checkpoint~\cite{Seesr} and do not perform additional fine-tuning on any clean or poisoned samples from the evaluated backdoor datasets. This keeps the restoration model independent of the evaluated backdoor attacks and prevents the purification process from relying on attack-specific data.}

\begin{table*}[t]
\centering
\caption{Defense performance comparison against classification backdoor attacks.}
\vspace{-3mm}
\setlength{\tabcolsep}{1mm}%
\resizebox{0.98\linewidth}{!}{%
\begin{tabular}{c|c|ccc|ccc|ccc|ccc|ccc|ccc|
>{\columncolor[HTML]{EFEFEF}}c
>{\columncolor[HTML]{EFEFEF}}c
>{\columncolor[HTML]{EFEFEF}}c}
\hline
 &  & \multicolumn{3}{c|}{\textbf{No Defense}} & \multicolumn{3}{c|}{\textbf{ShrinkPad}} & \multicolumn{3}{c|}{\textbf{BDMAE}} & \multicolumn{3}{c|}{\textbf{ZIP}} & \multicolumn{3}{c|}{\textbf{DataElixir}} & \multicolumn{3}{c|}{\textbf{SampDetox}} & \multicolumn{3}{c}{\cellcolor[HTML]{EFEFEF}\textbf{\methodName}} \\
\multirow{-2}{*}{Dataset} & \multirow{-2}{*}{Attack} & \textbf{CA$\uparrow$} & \textbf{ASR$\downarrow$} & \textbf{PA$\uparrow$} & \textbf{CA$\uparrow$} & \textbf{ASR$\downarrow$} & \textbf{PA$\uparrow$} & \textbf{CA$\uparrow$} & \textbf{ASR$\downarrow$} & \textbf{PA$\uparrow$} & \textbf{CA$\uparrow$} & \textbf{ASR$\downarrow$} & \textbf{PA$\uparrow$} & \textbf{CA$\uparrow$} & \textbf{ASR$\downarrow$} & \textbf{PA$\uparrow$} & \textbf{CA$\uparrow$} & \textbf{ASR$\downarrow$} & \textbf{PA$\uparrow$} & \textbf{CA$\uparrow$} & \textbf{ASR$\downarrow$} & \textbf{PA$\uparrow$} \\ \hline
 & No Attack & 88.48 & - & - & - & - & - & - & - & - & - & - & - & - & - & - & - & - & - & - & - & - \\
 & BadNet & 88.41 & 100.00 & 10.00 & 57.99 & 2.85 & 57.85 & \textbf{87.00} & \textbf{0.66} & 86.17 & 82.55 & 0.90 & 82.13 & 84.40 & 0.96 & 71.30 & 77.54 & 1.04 & 76.11 & 86.66 & 0.90 & \textbf{86.18} \\
 & PhysicalBA & 94.59 & 100.00 & 10.00 & 91.45 & 91.14 & 17.50 & 94.38 & 16.14 & 77.92 & 87.00 & 0.80 & 86.71 & \textbf{94.57} & 1.02 & 72.93 & 81.48 & 1.06 & 79.92 & 90.80 & \textbf{0.11} & \textbf{90.76} \\
 & Blended & 87.96 & 100.00 & 10.00 & 59.14 & \textbf{0.56} & 36.85 & 86.69 & 30.51 & 30.11 & 72.65 & 1.60 & 58.14 & \textbf{88.03} & 0.97 & 64.84 & 77.19 & 0.72 & 69.17 & 86.47 & 1.17 & \textbf{82.84} \\
 & Semantic Blended & 88.30 & 99.62 & 10.31 & 59.51 & 1.57 & 51.72 & 86.96 & 97.00 & 2.80 & 77.33 & 1.23 & 75.82 & \textbf{88.30} & \textbf{0.89} & 68.14 & 77.55 & 1.19 & 72.79 & 84.41 & 2.37 & \textbf{83.16} \\
 & WaNet & 88.47 & 99.84 & 10.10 & 58.41 & 13.22 & 53.52 & 86.33 & 57.27 & 32.58 & 58.77 & 9.12 & 55.79 & \textbf{87.55} & 2.77 & 67.51 & 76.89 & \textbf{0.51} & 74.27 & 83.11 & 1.41 & \textbf{82.08} \\
\multirow{-7}{*}{\begin{tabular}[c]{@{}c@{}}CIFAR-10\\ (10 classes)\end{tabular}} & Average & 89.36 & 99.89 & 10.08 & 65.30 & 21.86 & 43.48 & 88.27 & 40.32 & 45.92 & 75.66 & 2.73 & 71.71 & \textbf{88.57} & 1.32 & 68.94 & 78.13 & \textbf{0.90} & 74.45 & 86.29 & 1.19 & \textbf{85.00} \\ \hline
 & No Attack & 97.99 & - & - & - & - & - & - & - & - & - & - & - & - & - & - & - & - & - & - & - & - \\
 & BadNet & 98.31 & 100.00 & 5.70 & 41.84 & 3.30 & 41.78 & \textbf{97.50} & \textbf{0.30} & \textbf{97.72} & 89.85 & 3.63 & 88.34 & 94.38 & 2.49 & 41.73 & 54.83 & 4.06 & 53.95 & 93.44 & 0.39 & 92.77 \\
 & PhysicalBA & 97.88 & 100.00 & 5.70 & 43.87 & 3.04 & 43.67 & \textbf{96.95} & 0.34 & \textbf{97.03} & 90.07 & 0.80 & 89.92 & 94.31 & 1.18 & 39.14 & 52.23 & 1.93 & 51.72 & 93.73 & \textbf{0.26} & 93.01 \\
 & Blended & 98.17 & 100.00 & 5.70 & 42.94 & 3.58 & 31.13 & \textbf{97.67} & 17.67 & 54.98 & 78.85 & 3.42 & 66.56 & 94.26 & 2.28 & 31.24 & 52.66 & 2.82 & 42.00 & 93.69 & \textbf{0.15} & \textbf{84.56} \\
 & Semantic Blended & 98.03 & 98.95 & 6.68 & 65.80 & 29.31 & 46.93 & \textbf{97.31} & 98.20 & 1.77 & 79.33 & 11.55 & 68.27 & 96.16 & 6.04 & 33.91 & 51.63 & 10.41 & 43.90 & 91.55 & \textbf{5.27} & \textbf{80.96} \\
 & WaNet & 98.13 & 99.93 & 5.76 & 62.78 & \textbf{3.03} & 59.96 & \textbf{97.27} & 14.64 & 77.18 & 68.41 & 14.48 & 63.38 & 95.02 & 56.13 & 25.68 & 48.35 & 7.15 & 43.31 & 81.64 & 5.68 & \textbf{77.50} \\
\multirow{-7}{*}{\begin{tabular}[c]{@{}c@{}}GTSRB\\ (43 classes)\end{tabular}} & Average & 98.08 & 99.77 & 5.90 & 51.44 & 8.45 & 44.69 & \textbf{97.34} & 26.23 & 65.74 & 81.30 & 6.77 & 75.29 & 94.83 & 13.62 & 34.34 & 51.94 & 5.27 & 46.98 & 90.81 & \textbf{2.35} & \textbf{85.76} \\ \hline
 & No Attack & 86.70 & - & - & - & - & - & - & - & - & - & - & - & - & - & - & - & - & - & - & - & - \\
 & BadNet & 84.58 & 97.13 & 12.63 & 71.79 & 1.07 & 71.61 & 79.44 & 1.64 & 79.52 & 82.64 & 2.80 & 82.87 & \textbf{85.78} & \textbf{0.93} & \textbf{85.86} & 85.55 & 0.96 & 84.42 & 83.18 & 0.99 & 83.18 \\
 & PhysicalBA & 90.39 & 69.09 & 37.60 & 89.93 & 85.26 & 23.13 & 84.66 & 20.54 & 70.68 & 86.72 & 7.67 & 86.06 & \textbf{90.39} & 46.23 & 52.66 & 89.94 & 31.13 & 66.15 & 89.85 & \textbf{7.33} & \textbf{87.05} \\
 & Blended & 86.19 & 99.77 & 10.26 & 75.84 & 10.50 & 50.87 & 80.20 & 57.03 & 29.55 & 81.55 & 4.58 & 77.47 & \textbf{86.22} & 96.66 & 3.12 & 85.07 & 97.71 & 1.90 & 84.38 & \textbf{0.28} & \textbf{82.64} \\
 & Semantic Blended & 84.68 & 84.44 & 23.08 & 75.46 & 2.46 & 71.66 & 79.52 & \textbf{0.68} & 42.71 & 74.70 & 15.77 & 69.07 & \textbf{84.54} & 1.30 & 73.74 & 84.33 & 1.19 & 71.65 & 79.36 & 6.11 & \textbf{77.65} \\
 & WaNet & 84.38 & 99.18 & 10.54 & 75.89 & 2.20 & 75.41 & 79.49 & 98.42 & 1.50 & 80.68 & 2.26 & 79.43 & \textbf{84.31} & 26.74 & 63.26 & 83.36 & 16.20 & 71.11 & 81.80 & \textbf{1.21} & \textbf{82.34} \\
\multirow{-7}{*}{\begin{tabular}[c]{@{}c@{}}Imagenette\\ (10 classes)\end{tabular}} & Average & 86.15 & 89.92 & 18.82 & 77.78 & 20.29 & 58.53 & 80.66 & 35.66 & 44.79 & 81.25 & 6.61 & 78.79 & \textbf{86.25} & 34.37 & 55.73 & 85.65 & 29.44 & 59.05 & 83.71 & \textbf{3.18} & \textbf{82.57} \\ \hline
\end{tabular}
}
\vspace{-5mm}
\label{Baseline}
\end{table*}

\noindent\textbf{Evaluation Metrics for Classification Models.} We evaluate classification defenses using three metrics: (1) Clean Accuracy (\textit{CA} $\uparrow$), the percentage of purified clean images classified as their original labels by the poisoned model; (2) Attack Success Rate (\textit{ASR} $\downarrow$), the percentage of purified poisoned images classified as the attacker's target label; and (3) Poison Accuracy (\textit{PA} $\uparrow$), the percentage of purified poisoned images classified as their original labels by the poisoned model. We additionally report reference-based visual fidelity metrics between the original and purified images: Peak Signal-to-Noise Ratio (\textit{PSNR} $\uparrow$), Structural Similarity Index Measure (\textit{SSIM} $\uparrow$)~\cite{SSIM}, and Learned Perceptual Image Patch Similarity (\textit{LPIPS} $\downarrow$)~\cite{LPIPS}.

\noindent\textbf{Evaluation Metrics for CLIP-style Encoders.} We use the same CA, ASR, and PA metrics for CLIP-style encoders because the downstream few-shot image classification task follows the evaluation criteria as the classification setting.

{\noindent\textbf{Evaluation Metrics for Generative Models.} Following the image-to-image backdoor evaluation~\cite{I2IUAP}, we use SSIM~\cite{SSIM} to define two metrics: (1) Clean Similarity (\textit{CS} $\uparrow$), the SSIM between the output of the poisoned generative model given a purified clean low-resolution (LR) input and its corresponding high-resolution (HR) ground-truth image; and (2) Poison Similarity (\textit{PS} $\downarrow$), the SSIM between the output of the poisoned generative model given a purified poisoned LR input and the attacker's target image. Higher CS indicates better preservation of the intended generative output, whereas lower PS indicates stronger suppression of the backdoor target.}

\vspace{-2mm}
\subsection{Experimental Results}
\subsubsection{Performance Against Classification Backdoors}

{Table~\ref{Baseline} summarizes the performance of \methodName against classification backdoor attacks in comparison with ShrinkPad, BDMAE, ZIP, DataElixir, and SampDetox. \methodName consistently achieves a strong balance between backdoor mitigation and semantic information preservation. In particular, \methodName obtains the highest average PA on all three datasets, reaching 85.00\% on CIFAR-10, 85.76\% on GTSRB, and 82.57\% on Imagenette, while maintaining low average ASR values of 1.19\%, 2.35\%, and 3.18\%, respectively.}

{The magnitude of the PA improvement is especially evident under several attacks. For example, under Blended on CIFAR-10, \methodName improves PA by 19.8\% compared with the next-best method, increasing it from 69.17\% to 82.84\%, while its CA is only 1.56 points lower than the highest CA of 88.03\%. On GTSRB, \methodName improves PA under Blended by 27.0\% over ZIP, increasing it from 66.56\% to 84.56\%, while its CA is 3.98 points below the highest CA obtained by BDMAE. These results show that explicitly incorporating semantic guidance substantially improves the preservation of semantic information in inputs while maintaining competitive performance.
}

{The results also reveal different trade-offs among CA, ASR, and PA. DataElixir often achieves high CA because its selective detection strategy leaves samples that are not identified as poisoned unchanged. This better preserves clean inputs, but poisoned samples that are not detected can bypass purification, leading to higher ASR and lower PA in several settings. Similarly, BDMAE preserves high CA in many cases, but its average ASR reaches 40.32\%, 26.23\%, and 35.66\% on CIFAR-10, GTSRB, and Imagenette, respectively. In contrast, \methodName maintains consistently low ASR while achieving substantially higher PA across the three datasets.}

{For Semantic Blended, where the trigger itself contains semantically meaningful visual content, \methodName achieves the highest PA on all three datasets. Although DataElixir attains higher CA in these settings, \methodName preserves substantially more semantic information in poisoned samples while maintaining low ASR. This further shows that multi-level semantic guidance remains effective even when the backdoor trigger contains semantic content.}

\begin{table}[t]
\centering
\caption{{Performance comparison on preserving semantic details.}}
\vspace{-3mm}
\setlength{\tabcolsep}{0.8mm}%
\renewcommand{\arraystretch}{1.05}
\small

\begin{adjustbox}{max width=\linewidth}
\begin{tabular}{c|c|cccccc}
\hline
&  & \multicolumn{6}{c}{\textbf{Attack}} \\

\multirow{-2}{*}{Defense} &
\multirow{-2}{*}{Metric} &
\makecell{\textbf{No}\\[-1mm]\textbf{Attack}} &
\textbf{BadNet} &
\makecell{\textbf{Physical}\\[-1mm]\textbf{BA}} &
\textbf{Blended} &
\makecell{\textbf{Semantic}\\[-1mm]\textbf{Blended}} &
\textbf{WaNet} \\ \hline

& PSNR$\uparrow$
& $8.42{\pm}2.26$
& $8.41{\pm}2.35$
& $8.31{\pm}2.22$
& $8.56{\pm}1.96$
& $8.66{\pm}2.20$
& $8.26{\pm}2.38$ \\

& SSIM$\uparrow$
& $0.15{\pm}0.08$
& $0.15{\pm}0.09$
& $0.15{\pm}0.08$
& $0.05{\pm}0.01$
& $0.14{\pm}0.07$
& $0.15{\pm}0.08$ \\

\multirow{-3}{*}{ShrinkPad}
& LPIPS$\downarrow$
& $0.54{\pm}0.05$
& $0.54{\pm}0.06$
& $0.55{\pm}0.05$
& $0.69{\pm}0.05$
& $0.56{\pm}0.05$
& $0.56{\pm}0.04$ \\ \hline

& PSNR$\uparrow$
& $\mathbf{30.41{\pm}6.67}$
& $\mathbf{30.60{\pm}7.07}$
& $27.56{\pm}4.71$
& $\mathbf{26.74{\pm}1.86}$
& $\mathbf{29.32{\pm}4.23}$
& $\mathbf{40.41{\pm}6.93}$ \\

& SSIM$\uparrow$
& $\mathbf{0.93{\pm}0.05}$
& $\mathbf{0.93{\pm}0.05}$
& $\mathbf{0.90{\pm}0.05}$
& $\mathbf{0.82{\pm}0.02}$
& $\mathbf{0.92{\pm}0.04}$
& $\mathbf{0.99{\pm}0.01}$ \\

\multirow{-3}{*}{BDMAE}
& LPIPS$\downarrow$
& $\mathbf{0.05{\pm}0.04}$
& $\mathbf{0.05{\pm}0.04}$
& $0.08{\pm}0.04$
& $0.18{\pm}0.04$
& $\mathbf{0.07{\pm}0.04}$
& $\mathbf{0.01{\pm}0.01}$ \\ \hline

& PSNR$\uparrow$
& $24.81{\pm}2.91$
& $24.60{\pm}3.54$
& $24.40{\pm}3.29$
& $22.26{\pm}1.23$
& $19.95{\pm}2.13$
& $25.41{\pm}3.27$ \\

& SSIM$\uparrow$
& $0.70{\pm}0.13$
& $0.71{\pm}0.14$
& $0.71{\pm}0.12$
& $0.33{\pm}0.04$
& $0.51{\pm}0.15$
& $0.72{\pm}0.12$ \\

\multirow{-3}{*}{ZIP}
& LPIPS$\downarrow$
& $0.13{\pm}0.04$
& $0.13{\pm}0.05$
& $0.13{\pm}0.04$
& $0.47{\pm}0.08$
& $0.32{\pm}0.09$
& $0.14{\pm}0.04$ \\ \hline

& PSNR$\uparrow$
& $21.78{\pm}7.27$
& $21.56{\pm}7.12$
& $21.75{\pm}6.99$
& $18.24{\pm}4.19$
& $8.99{\pm}1.86$
& $21.92{\pm}7.01$ \\

& SSIM$\uparrow$
& $0.60{\pm}0.24$
& $0.60{\pm}0.24$
& $0.61{\pm}0.24$
& $0.22{\pm}0.10$
& $0.19{\pm}0.08$
& $0.61{\pm}0.23$ \\

\multirow{-3}{*}{DataElixir}
& LPIPS$\downarrow$
& $0.19{\pm}0.22$
& $0.19{\pm}0.22$
& $0.20{\pm}0.22$
& $0.24{\pm}0.21$
& $0.66{\pm}0.06$
& $0.20{\pm}0.21$ \\ \hline

& PSNR$\uparrow$
& $26.47{\pm}3.03$
& $26.19{\pm}2.88$
& $26.41{\pm}3.19$
& $21.06{\pm}0.56$
& $26.51{\pm}2.51$
& $26.65{\pm}2.77$ \\

& SSIM$\uparrow$
& $0.76{\pm}0.11$
& $0.76{\pm}0.11$
& $0.76{\pm}0.11$
& $0.40{\pm}0.06$
& $0.74{\pm}0.09$
& $0.76{\pm}0.10$ \\

\multirow{-3}{*}{SampDetox}
& LPIPS$\downarrow$
& $0.07{\pm}0.03$
& $0.07{\pm}0.03$
& $0.07{\pm}0.03$
& $\mathbf{0.12{\pm}0.03}$
& $0.09{\pm}0.03$
& $0.09{\pm}0.03$ \\ \hline

\rowcolor[HTML]{EFEFEF}
\cellcolor[HTML]{EFEFEF}
& PSNR$\uparrow$
& $28.09{\pm}2.91$
& $27.68{\pm}2.80$
& $\mathbf{27.63{\pm}2.57}$
& $23.32{\pm}0.79$
& $19.80{\pm}1.93$
& $27.56{\pm}2.84$ \\

\rowcolor[HTML]{EFEFEF}
\cellcolor[HTML]{EFEFEF}
& SSIM$\uparrow$
& $0.86{\pm}0.06$
& $0.86{\pm}0.06$
& $0.87{\pm}0.06$
& $0.45{\pm}0.08$
& $0.56{\pm}0.12$
& $0.86{\pm}0.05$ \\

\rowcolor[HTML]{EFEFEF}
\multirow{-3}{*}{
    \cellcolor[HTML]{EFEFEF}\textbf{\methodName}
}
& LPIPS$\downarrow$
& $\mathbf{0.05{\pm}0.03}$
& $0.06{\pm}0.04$
& $\mathbf{0.05{\pm}0.04}$
& $0.20{\pm}0.21$
& $0.24{\pm}0.07$
& $0.07{\pm}0.03$ \\ \hline

\end{tabular}
\end{adjustbox}

\vspace{-7mm}
\label{tab:visual_metrics}
\end{table}

{We evaluate the visual fidelity of purified images using PSNR, SSIM, and LPIPS, with the results summarized in Table~\ref{tab:visual_metrics}. BDMAE achieves the highest PSNR and SSIM in most settings. This is consistent with its localization-and-restoration mechanism, which selectively reconstructs regions identified as triggers while retaining the remaining input content. Consequently, when trigger localization is less effective, the purified image can remain highly similar to the poisoned input while also retaining the trigger. This behavior is reflected in Table~\ref{Baseline}, where BDMAE exhibits high average ASR values of 40.32\%, 26.23\%, and 35.66\% on CIFAR-10, GTSRB, and Imagenette, respectively.}

{In comparison, \methodName reduces the corresponding average ASR to 1.19\%, 2.35\%, and 3.18\%, while maintaining the highest average PA across the three datasets. Under PhysicalBA, for example, \methodName achieves an LPIPS of 0.05 compared with 0.08 for BDMAE, while also reducing Imagenette ASR from 20.54\% to 7.33\%. These results indicate that \methodName more effectively preserves semantic information relevant to downstream prediction while suppressing trigger-related information during purification.}

\begin{figure*}[t]
    \centering
    \includegraphics[width=0.80\linewidth]{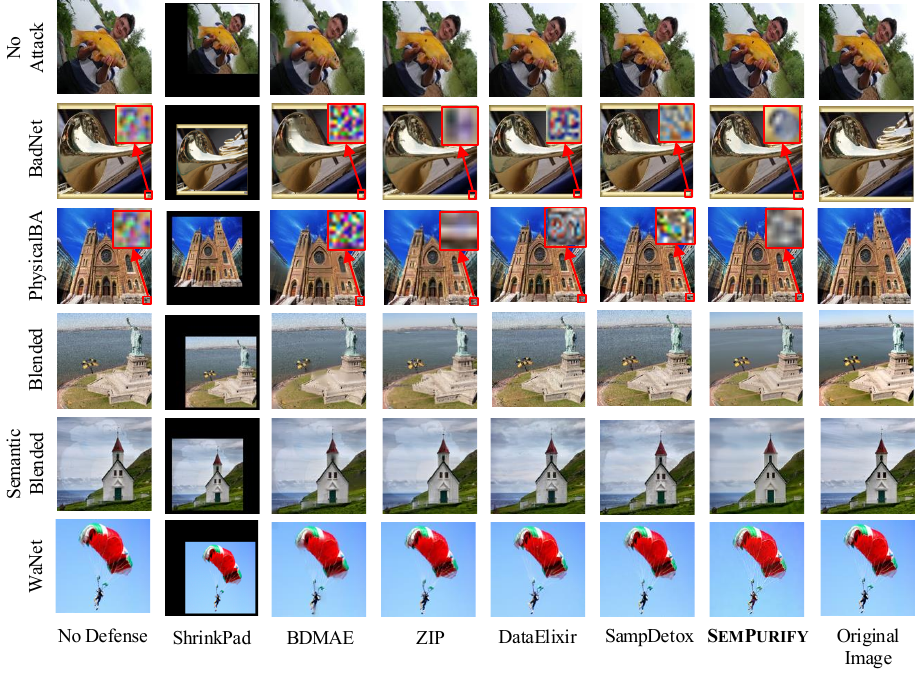}
    \vspace{-5mm}
    \caption{{Qualitative comparisons of images produced by different purification methods.}}
    \vspace{-5mm}
    \label{Visualfigures}
\end{figure*}

{Fig.~\ref{Visualfigures} provides representative qualitative comparisons of the purified images. The baseline methods generally preserve recognizable image content, but differences in fine-grained restoration and trigger removal can be observed. ShrinkPad introduces black regions due to its spatial transformation, while BDMAE largely preserves the original appearance but can retain visible trigger-related patterns in several examples. ZIP can alter foreground details under Blended, including people, parachutes, and statues. In the illustrated BadNet example, DataElixir alters facial details, and under Blended, residual global noise remains visible after SampDetox purification. In comparison, \methodName better suppresses the trigger while retaining the highlighted foreground structures and fine-grained details.}

\begin{table}[t]
\centering
\caption{Defense performance comparison against BadCLIP.}
\vspace{-3mm}

\small
\setlength{\tabcolsep}{1.3mm}
\renewcommand{\arraystretch}{1.05}

\resizebox{\linewidth}{!}{%
\begin{tabular}{
c|c|
c c c c c c|
>{\columncolor[HTML]{EFEFEF}}c
}
\hline
\textbf{Dataset}
& \textbf{Metric}
& \textbf{No Defense}
& \textbf{ShrinkPad}
& \textbf{BDMAE}
& \textbf{ZIP}
& \textbf{DataElixir}
& \textbf{SampDetox}
& \cellcolor[HTML]{EFEFEF}\textbf{\methodName}
\\
\hline

\multirow{3}{*}{
    \shortstack{Caltech101\\\cite{fei2004learning}}
}
& \textbf{CA$\uparrow$}
& 92.00
& \textbf{91.70}
& 86.24
& 90.70
& 90.39
& 90.61
& 91.30
\\

& \textbf{ASR$\downarrow$}
& 98.30
& 0.90
& 35.59
& 1.10
& 0.33
& 0.33
& \textbf{0.20}
\\

& \textbf{PA$\uparrow$}
& 2.60
& 91.00
& 57.97
& 90.70
& 91.27
& 91.16
& \textbf{91.30}
\\

\hline

\multirow{3}{*}{
    \shortstack{OxfordPets\\\cite{parkhi2012cats}}
}
& \textbf{CA$\uparrow$}
& 96.00
& 93.40
& 86.80
& 90.20
& \textbf{95.02}
& 89.26
& 93.40
\\

& \textbf{ASR$\downarrow$}
& 99.30
& 47.90
& 49.33
& 5.70
& 6.26
& 4.08
& \textbf{3.20}
\\

& \textbf{PA$\uparrow$}
& 5.50
& 51.30
& 48.99
& 90.40
& 91.44
& 92.11
& \textbf{93.30}
\\

\hline
\end{tabular}%
}

\vspace{-7mm}
\label{tab:caltech101_oxfordpets_results}
\end{table}

\vspace{-1mm}
\subsubsection{Performance Against CLIP Backdoors}

{Table~\ref{tab:caltech101_oxfordpets_results} summarizes the defense performance against BadCLIP. \methodName achieves the lowest ASR and the highest PA on both datasets, reaching an ASR of 0.20\% and PA of 91.30\% on Caltech101, and an ASR of 3.20\% and PA of 93.30\% on OxfordPets, while maintaining competitive CA.}

{The baselines exhibit different trade-offs between clean and poisoned inputs. On OxfordPets, ShrinkPad retains a CA of 93.40\% but leaves an ASR of 47.90\%, while ZIP reduces ASR to 5.70\% with a lower CA of 90.20\%. DataElixir achieves the highest CA of 95.02\% but has an ASR of 6.26\%, whereas SampDetox reaches an ASR of 4.08\% with a CA of 89.26\%. BDMAE exhibits substantially higher ASR on both Caltech101 and OxfordPets, reaching 35.59\% and 49.33\%, respectively, together with lower PA. In comparison, \methodName maintains competitive performance on clean inputs while consistently achieving the lowest ASR and highest PA, demonstrating the effectiveness of incorporating semantic guidance for CLIP-style representations.}

\begin{table}[t]
\centering
\caption{Defense performance comparison against generative backdoor attacks.}
\vspace{-3mm}
\setlength{\tabcolsep}{1mm}%
\resizebox{1\linewidth}{!}{%
\begin{tabular}{cccccccccc
>{\columncolor[HTML]{EFEFEF}}c
>{\columncolor[HTML]{EFEFEF}}c}
\hline
\multicolumn{2}{c}{\textbf{No Defense}}
& \multicolumn{2}{c}{\textbf{BDMAE}}
& \multicolumn{2}{c}{\textbf{ZIP}}
& \multicolumn{2}{c}{\textbf{DataElixir}}
& \multicolumn{2}{c}{\textbf{SampDetox}}
& \multicolumn{2}{c}{\cellcolor[HTML]{EFEFEF}\textbf{\methodName}} \\
\textbf{CS$\uparrow$} & \textbf{PS$\downarrow$}
& \textbf{CS$\uparrow$} & \textbf{PS$\downarrow$}
& \textbf{CS$\uparrow$} & \textbf{PS$\downarrow$}
& \textbf{CS$\uparrow$} & \textbf{PS$\downarrow$}
& \textbf{CS$\uparrow$} & \textbf{PS$\downarrow$}
& \textbf{CS$\uparrow$} & \textbf{PS$\downarrow$} \\ \hline
\textbf{0.89} & 0.86
& 0.54 & 0.24
& 0.69 & 0.38
& 0.61 & 0.21
& 0.66 & 0.27
& 0.78 & \textbf{0.20} \\ \hline
\end{tabular}%
}
\label{tab:I2I Purificationadditional}
\end{table}

\begin{figure}[t]
    \centering
    \includegraphics[width=0.85\linewidth]{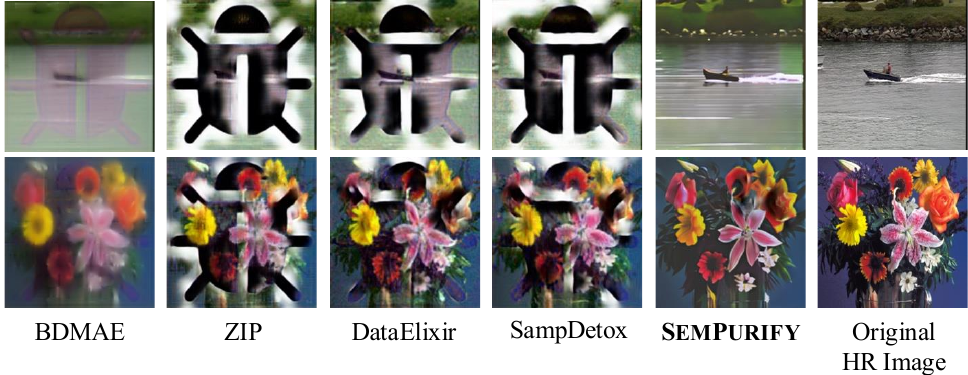}
    \vspace{-3mm}
    \caption{Comparison of the generated image using different purification methods.}
    \vspace{-6mm}
    \label{fig:I2I backdoor comparison}
\end{figure}

\subsubsection{Performance Against Generative Backdoors}

For generative backdoor attacks, we compare \methodName with {BDMAE,} ZIP, DataElixir, and SampDetox. ShrinkPad is omitted due to its suboptimal performance in this setting. 

{As shown in Table~\ref{tab:I2I Purificationadditional}, \methodName achieves both the highest CS and the lowest PS among the evaluated defenses. Specifically, it reduces PS from 0.86 without defense to 0.20 while maintaining a CS of 0.78. BDMAE and DataElixir also reduce PS to 0.24 and 0.21, respectively, but their CS decreases substantially to 0.54 and 0.61. ZIP and SampDetox retain higher CS values of 0.69 and 0.66, but yield higher PS values of 0.38 and 0.27. These results show that \methodName provides a stronger balance between suppressing the targeted backdoor output and preserving semantic information in the generated result.}

{Fig.~\ref{fig:I2I backdoor comparison} provides representative outputs generated from purified poisoned inputs, where the attack target is a black bug pattern. In the illustrated examples, the target pattern is substantially suppressed after \methodName purification, whereas it remains visible after the other purification methods. This qualitative observation is consistent with the low PS achieved by \methodName in Table~\ref{tab:I2I Purificationadditional}.}

\vspace{-2mm}
\subsection{Impact of Semantic Embeddings and Tags}
\vspace{-2mm}

We investigate the impact of semantic embeddings ($S_e$) and image tags ($S_t$) on the performance of \methodName. Specifically, we evaluate \methodName's performance in classification backdoor attacks on Imagenette using only $S_e$, only $S_t$, and both. The results are presented in Table~\ref{Ablation study of embeddings and Tags}. 

{The ablation results suggest that the two forms of semantic guidance play different but complementary roles during restoration. $S_e$ provides feature-level information that helps anchor the restoration process to the detailed visual content of the original image, whereas $S_t$ provides a higher-level description of the prominent objects and scene semantics. As a result, the restoration process is guided not only by detailed visual representations, but also by a more abstract semantic description of what content should be preserved. This multi-level guidance helps reduce the dependence on any single semantic representation and provides a more complete description of the original input.}

{The benefit of combining the two forms of guidance is particularly evident under several challenging attacks, including PhysicalBA, Blended, and Semantic Blended, where using both $S_e$ and $S_t$ provides a better overall trade-off between preserving benign image information and suppressing the backdoor effect. For example, under Semantic Blended, incorporating $S_t$ together with $S_e$ increases PA from 72.58\% to 77.65\% while reducing ASR from 6.96\% to 6.11\%. This result is especially meaningful because Semantic Blended introduces semantically meaningful trigger content, for which high-level semantic guidance can provide additional constraints during restoration.}

\begin{table}[t]
\centering
\caption{Impact of semantic embeddings and image tags.}
\vspace{-3mm}
\setlength{\tabcolsep}{1mm}%
\resizebox{0.80\linewidth}{!}{%
\begin{tabular}{c|ccc|ccc|
>{\columncolor[HTML]{EFEFEF}}c 
>{\columncolor[HTML]{EFEFEF}}c 
>{\columncolor[HTML]{EFEFEF}}c }
\hline
 & \multicolumn{3}{c|}{\textbf{w/ $S_e$}} 
 & \multicolumn{3}{c|}{\textbf{w/ $S_t$}} 
 & \multicolumn{3}{c}{\cellcolor[HTML]{EFEFEF}\textbf{w/ $S_e + S_t$}} \\
\multirow{-2}{*}{Attack} 
& \textbf{CA$\uparrow$} & \textbf{ASR $\downarrow$} & \textbf{PA $\uparrow$} 
& \textbf{CA $\uparrow$} & \textbf{ASR$\downarrow$} & \textbf{PA$\uparrow$} 
& \textbf{CA $\uparrow$} & \textbf{ASR$\downarrow$} & \textbf{PA $\uparrow$} \\ \hline
BadNet & \textbf{83.21} & 1.07 & \textbf{83.36} & 68.84 & \textbf{0.16} & 69.22 & 83.18 & 0.99 & 83.18 \\
PhysicalBA & 89.63 & 7.36 & 86.85 & 77.50 & \textbf{3.6} & 77.09 & \textbf{89.85} & 7.33 & \textbf{87.05} \\
Blended & 84.02 & 0.31 & 81.98 & 70.92 & \textbf{0.02} & 70.77 & \textbf{84.38} & 0.28 & \textbf{82.64} \\
Semantic Blended & 73.42 & 6.96 & 72.58 & 56.78 & \textbf{0.31} & 60.22 & \textbf{79.36} & 6.11 & \textbf{77.65} \\
WaNet & \textbf{82.39} & 1.80 & \textbf{82.70} & 65.75 & \textbf{0.39} & 65.37 & 76.94 & 1.55 & 76.91 \\ \hline
\end{tabular}
}
\vspace{-3mm}
\label{Ablation study of embeddings and Tags}
\end{table}

\vspace{-2mm}
\subsection{Inference Efficiency}
\vspace{-2mm}

{To assess the practicality of \methodName for real-world deployment, we evaluate its efficiency on 50 Imagenette images at $256\times256$ resolution using a single NVIDIA A100 GPU. As shown in Table~\ref{tab:efficiency}, \methodName processes each image in 0.5101 seconds, which is faster than ZIP (1.0441s), DataElixir (15.2913s), and SampDetox (12.2146s), although ShrinkPad and BDMAE are more lightweight. \methodName incurs a higher peak GPU memory usage of 5764.6 MB due to the use of a large pre-trained T2I model. Nevertheless, this memory overhead remains acceptable in practice, while the sub-second inference time makes the overall computational cost suitable for deployment.}

\begin{table}[t]
\centering
\caption{{Computational cost on Imagenette.}}
\vspace{-3mm}
\setlength{\tabcolsep}{1.2mm}%
\resizebox{\linewidth}{!}{%
\begin{tabular}{c|cccccc}
\hline
\textbf{Metric} & \textbf{ShrinkPad} & \textbf{ZIP} & \textbf{BDMAE} & \textbf{DataElixir} & \textbf{SampDetox} & \cellcolor[HTML]{EFEFEF}\textbf{\methodName} \\ \hline
Latency (Second)$\downarrow$      & \textbf{0.0047} & 1.0441 & 0.2085 & 15.2913 & 12.2146 & \cellcolor[HTML]{EFEFEF}0.5101 \\
Peak GPU Mem (MB)$\downarrow$    & \textbf{100.7}  & 1579.3 & 451.5  & 1632.1  & 1737.1  & \cellcolor[HTML]{EFEFEF}5764.6 \\ \hline
\end{tabular}
}
\vspace{-7mm}
\label{tab:efficiency}
\end{table}

\vspace{-2mm}
\section{Conclusion}
\vspace{-2mm}

{In this paper, we propose \methodName, a semantics-aware purification framework that incorporates semantic embeddings and tag-level prompts into a T2I model to remove backdoor triggers while preserving semantic information. Extensive experiments across classification models, CLIP-style encoders, and generative models demonstrate that \methodName effectively suppresses backdoor behavior while maintaining high PA and preserving semantic information across diverse attacks and datasets. These results highlight the benefit of explicitly incorporating semantic guidance into black-box backdoor purification.}


%
%
\bibliographystyle{splncs04}
\bibliography{main}
\end{document}